\documentclass[sn-mathphys,Numbered]{sn-jnl}

\usepackage{graphicx}%
\usepackage{multirow}%
\usepackage{amsmath,amssymb,amsfonts}%
\usepackage{amsthm}%
\usepackage{mathrsfs}%
\usepackage[title]{appendix}%
\usepackage{xcolor}%
\usepackage{textcomp}%
\usepackage{manyfoot}%
\usepackage{booktabs}%
\usepackage{algorithm}%
\usepackage{algorithmicx}%
\usepackage{algpseudocode}%
\usepackage{listings}%

\usepackage{tabularx}
\usepackage[whole]{bxcjkjatype} 
\usepackage{enumitem}
\usepackage{afterpage}
\usepackage{stfloats}
\usepackage{fancyhdr}
\usepackage{caption}
\usepackage{lmodern}
\usepackage{anyfontsize}

\begin{document}

\title{Japanese Tort-case Dataset for Rationale-supported Legal Judgment Prediction}


\author*[1]{\fnm{Hiroaki} \sur{Yamada}}\email{yamada@c.titech.ac.jp}

\author[1]{\fnm{Takenobu} \sur{Tokunaga}}\email{take@c.titech.ac.jp}

\author[2,3]{\fnm{Ryutaro} \sur{Ohara}}\email{r.ohara@ntmlo.com}

\author[3]{\fnm{Akira} \sur{Tokutsu}}\email{a.tokutsu@r.hit-u.ac.jp}

\author[3]{\fnm{Keisuke} \sur{Takeshita}}\email{kei.takeshita@r.hit-u.ac.jp}

\author[3]{\fnm{Mihoko} \sur{Sumida}}\email{m.sumida@r.hit-u.ac.jp}

\affil[1]{\orgname{Tokyo Institute of Technology}, \state{Tokyo}, \country{Japan}}

\affil[2]{\orgname{Nakamura, Tsunoda \& Matsumoto}, \state{Tokyo}, \country{Japan}}

\affil[3]{\orgname{Hitotsubashi University}, \state{Tokyo}, \country{Japan}}


\abstract{This paper presents the first dataset for Japanese Legal Judgment Prediction (LJP), the Japanese Tort-case Dataset (JTD), 
which features two tasks: tort prediction and its rationale extraction. The rationale extraction task identifies the court's accepting arguments from alleged arguments by plaintiffs and defendants, which is a novel task in the field.
JTD is constructed based on annotated 3,477 Japanese Civil Code judgments by 41 legal experts, resulting in 7,978 instances with 59,697 of their alleged arguments from the involved parties.
Our baseline experiments show the feasibility of the proposed two tasks, and our error analysis by legal experts identifies sources of errors and suggests future directions of the LJP research.}




\maketitle

\section{Introduction}
Legal information processing aims to provide computational aid in legal procedures, including predicting the outcome of a case through legal judgment prediction (LJP).
LJP is beneficial to both legal professionals and the general public.
It allows them to predict litigation outcomes in advance, and they can take better actions based on expectations. For example, they can get faster conciliation and smoother negotiations, resulting in more efficient legal services.
LJP also reduces the cost of legal services and improves their access. Easier access to justice is important for those with limited or no access to traditional legal services.

LJP has been a longstanding research topic in artificial intelligence and, like other domains, has adopted machine learning (ML) techniques.
The ML techniques require large datasets for training and evaluating ML models.
\citet{Xiao2018-nc} proposed a dataset of 2.6M Chinese criminal cases annotated with applicable laws, charges, and prison terms. \citet{Chalkidis2019-dz} presented a dataset of 11.5K cases from the European Court of Human Rights, which is designed for violated article detection and case importance prediction. \citet{Katz2017-dl} constructed a dataset of 28K cases from the Supreme Court of the United States. \citet{semo-etal-2022-classactionprediction} released the LJP dataset focused on class action cases in the United States.
\citet{DBLP:journals/corr/abs-2110-00976} proposed a collection of datasets to evaluate model performance across different legal tasks, including LJP tasks in English.
In contrast, there is no LJP dataset employing real judgment documents in the Japanese jurisdiction.
The LJP tasks and their datasets should be designed to reflect differences in jurisdictions.
Against this backdrop, we construct the first dataset of LJP,
\textbf{Japanese Tort-case Dataset (JTD)},
for the Japanese LJP research.

In JTD, we deal with judgment on civil cases about torts~(\href{https://www.japaneselawtranslation.go.jp/en/laws/view/3494\#je_pt3ch5at1}{Civil Code, Art.~709})\footnote{\citet{BB29009904} provides a good overview of the Japanese torts.}.
Tort is an important and popular topic in civil cases.
Japanese law affirms a tort as a negligent or intentional infringement of rights or legal interests that cause a plaintiff to suffer loss or harm.
In modern society, torts play an important role in disputes on the internet, for example, cases of defamation and privacy infringement on social media. In such cases, tort law is often used to determine liability since there is usually no explicit contract between the parties.

\begin{figure}[t]
\centering
\includegraphics[width=1.0\linewidth]{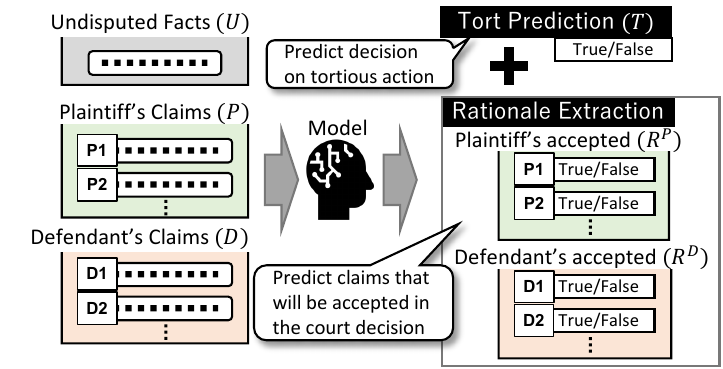}
\caption{Rationale-supported legal judgment prediction featuring a tort case}
\label{fig:general_framework}
\end{figure}
\begin{figure}[ht]
\centering
\includegraphics[width=0.95\linewidth]{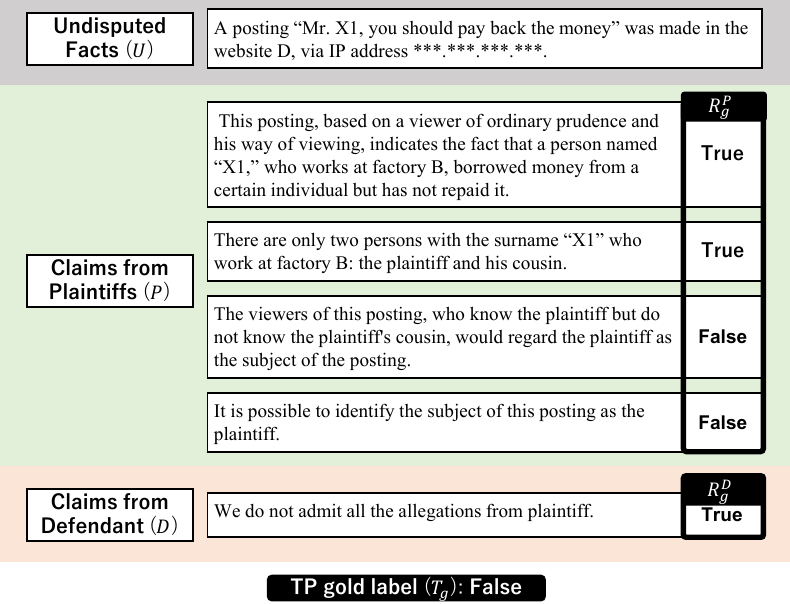}
\caption{An example about a defamation case. Translations are ours and modified for presentation purpose. $R_{g}^{P}$ and $R_{g}^{D}$ represent gold labels for $R^{P}$ and $R^{D}$, respectively.}
\label{fig:dataset_sample_simple}
\end{figure}

Figure~\ref{fig:general_framework} shows an overview of our two tasks: \textbf{Tort Prediction (TP)} and \textbf{Rationale Extraction (RE)}.
A tort case involves two parties: plaintiffs and defendants. Plaintiffs are claimants of the case, arguing that a defendant's action is a tort, while defendants contest plaintiffs' arguments.
TP predicts whether a tort is affirmed ($T$, a Boolean value), given undisputed facts ($U$) and arguments from both parties ($P$ from plaintiffs and $D$ from defendants).
Undisputed facts are not disputed by any parties or agreed upon by both parties. 
They provide the LJP model with context to validate the parties' arguments.
The final decision on a tort ($T$) should be based on the arguments that are accepted by the judge.
Thus the accepted arguments can be considered rationales for the final decision ($T$). RE identifies the accepted arguments ($R^P$ for plaintiffs and $R^D$ for defendants, both are sequences of Boolean values, denoting accepted arguments as True) in the parties' arguments ($P$ and $D$).
To summarise, our tasks take $(U, P, D)$ as input and output $(T, R^{P}, R^{D})$.

Figure~\ref{fig:dataset_sample_simple} shows an example of an instance.
There is an undisputed fact ($U$), four claims from the plaintiff ($P$), and one claim from the defendant ($D$).
A gold standard for the tort prediction task is false ($T_{gold}$), meaning the subject of this instance is not considered a tort.
Gold standard labels for the rationale extraction task are $\{True, True, False, False\}$ for the plaintiff ($R^{P}_{g}$) and $\{False\}$ for the defendant~($R^{D}_{g}$).

Our main contributions are the following.
We propose new tasks for the Japanese LJP, which consist of judicial decision prediction and identification of their rationales.
We conducted a large-scale annotation with 41 legal experts. In the annotation, the annotators captured direct causal relations between the court decisions and arguments from the parties, allowing multiple court decisions on multiple subject matters in a case.
From the 3,477 annotated documents, we built JTD consisting of 7,978 tort-related instances.
To establish baseline performance with the dataset, we conduct experiments employing hierarchical Transformer architecture and multi-task learning approaches. Moreover, we perform a detailed error analysis by legal experts to identify sources of errors and suggest future research directions.
Our dataset will be available for researchers on a website.

\section{Related Work and Background}
ML-based approaches have been popular in LJP research.
ML-based systems automatically learn how judges make decisions from a large number of cases (e.g., judgment documents).
These models take fact descriptions as input and predict outcomes or relevant laws. European Court of Human Rights cases are popular sources of datasets for LJP~\citep{Aletras2016-fc,Medvedeva_undated-jd,Chalkidis2019-dz, 10.1162/tacl_a_00532}. 
\citet{DBLP:conf/jurix/GalliGFGLPRST22} constructed a dataset for an outcome prediction task for Italian Value Added Tax decisions.
\citet{Katz2017-dl} built a dataset of cases from the Supreme Court of the United States to train their models.
\citet{semo-etal-2022-classactionprediction} constructed another LJP dataset of class action cases in the United States. 
LJP on Chinese Criminal cases is another big venue of ML-based LJP models~\citep{Luo2017-ar,Zhong2018-lb,Hu2018-rz,Long2019-li,Xu2020-un}.
A Japanese dataset for legal tasks is available for Competition on Legal Information Extraction/Entailment (COLIEE)~\citep{DBLP:conf/jsai/RabeloKGYKS20}. However, COLIEE is designed for legal entailment and information retrieval on the Japanese bar exam, and its data size is limited.

Due to the lack of a large reliable dataset, the Japanese LJP research has hardly employed the ML-based approach.
Instead, the symbolic approach has been popular for Japanese LJP. The symbolic systems predict outcomes of legal reasoning by rules and logic~\cite{Nitta1993-vx,Nitta1993-qk}.
Although the symbolic systems require human experts’ intervention in development, we can easily interpret their behaviour.
PROLEG~\cite{Satoh2011-cz} demonstrated that a logic programming system could work for the Japanese legal system. PROLEG is a legal reasoning system based on Prolog,  implementing a decision-making theory used in civil litigation in Japan.
However, there still needs to be a solution to extracting logical clauses from natural language text~\cite{Navas-Loro2019-sr}.

Moreover, in the recent era of rapid socio-economic change, the importance of a type of provision known as \textit{general clause} is growing in legal practice; the general clauses enable legislators to deal with various unforeseeable situations and to apply the law fairly and appropriately.
They do not specify the physical or social facts that must be proven to determine whether specific legal requirements have been met. Instead, they only provide abstractive concepts as requirements.
The fulfilment of those requirements will be determined through the comprehensive assessment or evaluation of the relevant individual facts in concrete cases.
Determining their fulfilment can hardly be implemented by rule-based or logic programming-based approaches.
On the other hand, the ML-based approach can learn the standards of those requirements from many precedents and perform better.
Therefore, it is essential to construct a large-scale dataset of Japanese judgment documents to facilitate the ML-based approach for Japanese LJP tasks.
We use real Civil Code judgment documents, specifically, torts cases, since the basic rule of torts in the Japanese Civil Code (\href{https://www.japaneselawtranslation.go.jp/en/laws/view/3494\#je_pt3ch5at1}{Art.~709}) is a good example of the above-mentioned general clause.

The success of deep learning methods in many areas raises concerns about the lack of explainability~\citep{Jacovi2020-bd} behind their output.
LJP outputs are expected to align with \textit{Justice}, and LJP systems should be trusted by society. Therefore, explainability is more important and serious in the legal domain than in other domains.
Even if LJP systems are used as assistant tools for legal consulting, they can affect people's behaviours and indirectly influence their social status and assets. Thus, the LJP system has to explain the reason for predictions.
To accommodate the needs, recent LJP studies introduced explanation tasks, including court view generation~\cite{Ye2018-ud}, rationale paragraph extraction~\cite{Chalkidis2021-bh}, and case features extraction as rationales~\cite{Ferro2019-cu,Branting2021-io}.
Following the prior work, we design our explanation task as rationale extraction similar to \citet{Chalkidis2021-bh}, but our extraction task is at span level instead of paragraph level. In addition, our target of rationale extraction is argumentative claims from the parties (e.g., plaintiff's factual allegations) instead of submitted fact descriptions.

\section{Japanese Tort-case Dataset (JTD)}
\subsection{Data Source\label{sec:dataset_source}}
Our data source of the judgment documents is a legal database ``Hanreihisho\footnote{\url{https://www.hanreihisho.com/}}'' provided by LIC Co., Ltd. 
We curated judgment documents from the first instances of Civil Code cases from lower courts.
We retrieved the documents from the database using keyword-based queries shown in Table~\ref{tab:queries}.
We retrieved documents that describe general tort cases of defamation, privacy infringement and reputation injury using query A.
As tort disputes on the Internet are often discussed in Disclosure of Identification Information of the Sender (DIIS) cases, we also retrieved DIIS cases using query B.
\begin{table}[t]
\centering
\small
\caption{Queries used in our document retrieval. Translations in square brackets are ours.}
\begin{tabular}{@{}lp{0.8\linewidth}@{}}
\toprule
\textbf{Type} &
  \textbf{Query} \\ \midrule
Query A &
  (``名誉 [fame]'' OR ``プライバシー [privacy]'' OR ``信用毀損 [damage to credibility]'' ) AND ``不法行為 [tort]''  NOT ``発信者情報開示 [DIIS]'' NOT ``地位確認 [status confirmation]'' NOT ``無効確認 [declaration of nullity]''  NOT ``商標 [trademark]'' \\
Query B &
  ``発信者情報開示 (DIIS)'' \\ \bottomrule
\end{tabular}
    \label{tab:queries}
\end{table}
As they might include non-tort cases,
we manually excluded the irrelevant documents later in the human annotation process.

When writing a judgment document, the judges often comply with a particular guideline for writing civil case judgments~\cite{JRTI-j-2020}. This leads to high similarities in structure across judgments.
The structure starts with \textit{Main Text} briefly describing the final decision, followed by 
\textit{Facts and Reasons} containing sections: a summary of the case, undisputed facts, arguments from the parties, and the court's decisions (detailed judicial decisions).

As our target documents describe court cases, the documents can contain personal information, sensitive information of parties or legally protected information such as trade secrets.
\citet{Leins2020-ak} sheds light on potential ethical issues in constructing datasets from a sensitive data source like judgment documents.
In the Japanese legal system, it is guaranteed that anyone can access judgment documents by law (\href{https://www.japaneselawtranslation.go.jp/en/laws/view/2834\#je_pt1ch5sc1at5}{Code of Civil Procedure, Art.~91}), but the parties may request to opt out of giving others access to their judgment documents (\href{https://www.japaneselawtranslation.go.jp/en/laws/view/2834\#je_pt1ch5sc1at6}{Code of Civil Procedure, Art.~92}).
Therefore, sensitive secrets should not be contained in the database we use.
Moreover, the providers of documents and databases pseudonymise the documents before publishing a case in journals or databases.

\subsection{Annotation Scheme}
To obtain a set of tuples ($U, P, D, T_{g}, R^{P}_{g}, R^{D}_{g}$) for our task inputs and outputs, we annotate the following information based on the work of \citet{yamada-etal-2022-annotation}.
Here, subscripts $g$ denote task answers (golds).
Annotators extract spans at the character level according to the following definitions below.
Annotators may not identify any spans for a type if there is no corresponding text in a document.

The \textbf{Court Decisions (CD)} span describes a judge's decision on tort, which should be found in the judicial decision section.
The CD span has a \textbf{Decision (@D)} attribute, indicating whether judges affirmed the tort (True) or not (False).

The \textbf{Claim (CL)} span describes important parties' claims which are relevant to the decision on torts\footnote{In reality, we annotated two subcategories of CL: Factual Claims (FC) and Claims of Norms (NC); the former includes factual allegations and their opposing fact assertions, while the latter refers to abstract legal arguments regarding torts (e.g., references to precedents from the supreme court).}.
The CL span has two attributes: \textbf{Accepted (@AC)} indicating whether judges accepted the claim (True) or not (False), and \textbf{Who (@W)} indicating their claimant, i.e. one of \textit{Plaintiff}, \textit{Defendant} and \textit{Other}\footnote{We did not consider \textit{Other} in our later experiments as there were few.} (e.g., a third party with interest in the outcome).

The \textbf{Undisputed Facts (UF)} fact describes facts that are undisputed by any parties. 
The UF spans should be identified in sections other than the judicial decision. The UF span has no attribute.

A judgment document might have multiple CD spans.
Note that CD spans in a document can have different values since they concern different decisions, though the decisions can be related to each other.
The annotators associate each non-CD annotated span (CL and UF spans) with its relevant CD span.
A non-CD annotated span can be associated with multiple CD spans.

Our annotation scheme follows \citet{Chalkidis2021-bh}, but with more precise annotation granularity. First, we aim to annotate each individual subject (CD span) on trial instead of annotating an entire judgment document, which is a more precise task design in terms of simulating judicial decision-making. Also, our scheme allows annotators to extract rationale spans at the character level instead of the paragraph level.
Moreover, our scheme captures argumentative claims from the parties (factual allegations) in addition to submitted fact descriptions. 
While the annotation of the argumentative claims was implemented in previous work~\citep{DBLP:conf/jurix/GalliGFGLPRST22}, ours additionally implements labels indicating whether they are accepted by the court or not.

\subsection{Annotation Procedure}
Annotators get 16 pages of guidelines and five sample annotated documents with commentaries.
The guidelines are available as a part of our dataset package.
First, the annotators read through a judgment document to understand the argument flow.
They discard the document if it does not concern torts.
After the screening, the annotators annotate spans (CD, CL and UF) and assign necessary attributes to each span.
Because the information of CL and UF spans are used as inputs to a model, they are not annotated in the judicial decision section.
The annotators may refer to the judicial decision section only for assigning attribute @D to CD and @AC to CL.
Finally, the annotators associate every non-CD annotated span with its corresponding CD span.
To balance the workload between annotators, we let annotators skip a document which contains more than 15 CD spans.

\subsection{Annotation Study}
We assessed the reproducibility of the annotation scheme with five annotators, including three lawyers, one law school graduate and one undergraduate.
We asked them to annotate 25 documents independently.

As our tasks identify appropriate units in text, which are meaningful and annotatable, 
in other words, ``unitising'' tasks, we use Krippendorff's $\alpha_U$~\cite{Krippendorff1995-mf} as the main metric for the inter-annotator agreement (IAA)\footnote{
We used the implementation by \citet{Meyer2014-er}.}. We calculate  $\alpha_U$ using the character offset of spans and their ``labels''.
The span types\footnote{Four span types: CD, UF, NC and FC. Note that NC and FC are the subcategories of CL.} are used as a label for the spans, while
the attribute values are a label for the attributes.
In calculating $\alpha_U$ for the span association, a label for each non-CD annotated span is its associated CD span.
As the boundaries of a CD span might be different between annotators, we merged overlapped spans from different annotators into a single CD span taking the union of them.

We observed good agreement overall. 
The annotators identified $377.4$ spans on average from the 25 documents, achieving $0.654$ of $\alpha_U$ over all span types.
The $\alpha_U$ of attributes are at $0.629$ (@AC), $0.641$ (@W) and $0.608$ (@D) showing reproducible anntotions. 
$\alpha_U$ of span association shows a reasonable score at $0.430$, but it is still lower than that of span extraction due to the error propagation from the span extraction.

\subsection{Production Annotation}
To increase instances, we deployed the annotation scheme to more annotators with legal knowledge and experience. Each document was annotated by an annotator.
We qualified annotators through dry-run annotations to maintain annotation quality. 
As a result, 41 annotators participated in the production annotation.
The annotators consist of nine professional lawyers, 22 law school graduates, and 10 undergraduates in law.
Graduating from law school or passing the preliminary exam is a prerequisite for taking the Japanese Bar exam.
All the undergraduates were going to take the Bar exam, and 8 of them had already passed the preliminary exam when the annotation had started.
We also checked whether the annotator complied with the annotation guideline in the middle of and after the annotation period and excluded severe violators. 
During their annotation, the annotators could ask questions and discuss edge cases with other annotators via a text-based communication workspace.

\subsection{Dataset Construction\label{sec:dataset_construction}}
\begin{figure}[t]
\centering
\includegraphics[width=0.75\linewidth]{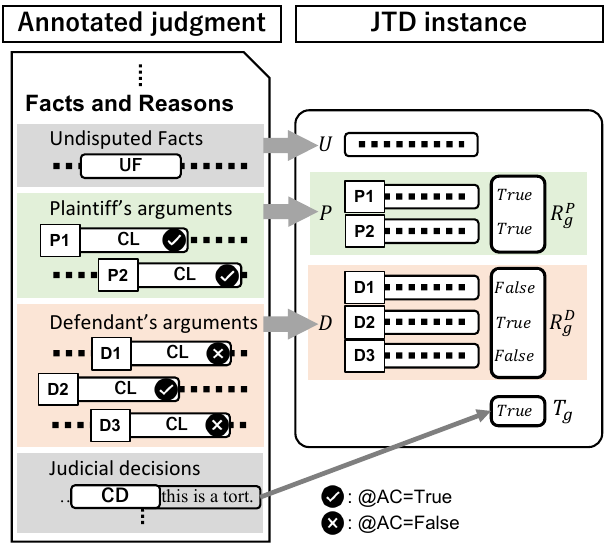}
\caption{Dataset Construction}
\label{fig:dataset_construction}
\end{figure}
We construct the JTD from the annotated judgment documents (Figure~\ref{fig:dataset_construction}). JTD consists of a set of tuples ($U,P,D,T_g,R^P_g,R^D_g$); we call each an instance.
$U$ is made by text annotated as UF.
$P$ and $D$ are sequences of text, which are annotated as CL, from the plaintiffs and the defendants, respectively.
Their corresponding gold labels for the RE task are $R^P_g$ and $R^D_g$.
Each of them is a sequence of Boolean flags imported from attributes @AC.
The orders of elements in those sequences correspond to their appearance in the judgment document.
$T_g$ is a single Boolean value made by attribute @D which is annotated on a CD span.

\begin{table}[t]
\caption{Japanese Tort-case Dataset overview\label{tab:dataset_overview}}
\centering
\small
\begin{tabular}{@{}lr@{}}
\toprule
\# of docs              & 3,477  \\
avg. instances/doc          & 2.3   \\ \midrule
\# of instances             & 7,978  \\
\# of Claims & 59,697 \\
\# of Undisputed Facts & 10,236 \\
\bottomrule
\end{tabular}
\end{table}

\begin{table}[t]
\caption{Dataset split. We split the dataset according to the number of instances.\label{tab:data_split}}
\centering
\small
\begin{tabular}{@{}lrrr@{}}
\toprule
Split & \# of docs &\# of instances  &  \# of claims\\
\midrule
Dev   & 329  & 803 & 6,063\\
Test   & 391 & 811 & 5,945\\
Train  & 2,757 & 6,364  & 47,689\\
\midrule
All  & 3,477 & 7,978 & 59,697 \\
\bottomrule
\end{tabular}
\end{table}

\begin{table}[ht]
\caption{Label (tort or not) distribution of instances.\label{tab:stat_tort}}
\centering
\small
\begin{tabular}{@{}lrrrr@{}}
\toprule
 & \multicolumn{1}{c}{True} & \multicolumn{1}{c}{False} & \multicolumn{1}{c}{All} & \multicolumn{1}{c}{True rate} \\
\midrule
Dev   & 304  & 499  & 803  & 37.9\% \\
Test  & 381  & 430  & 811  & 47.0\% \\
Train & 2,488 & 3,876 & 6,364 & 39.1\% \\
\midrule
All   & 3,173 & 4,805 & 7,978 & 39.8\% \\
\bottomrule
\end{tabular}
\end{table}

\begin{table}[ht]
\caption{Label (accepted or not) distribution of claims.\label{tab:stat_claims}}
\centering
\small
\begin{tabular}{@{}lrrrr@{}}
\toprule
      & \multicolumn{4}{c}{Overall}         \\
\cmidrule(rl){2-5}
 & \multicolumn{1}{c}{True} & \multicolumn{1}{c}{False} & \multicolumn{1}{c}{All} & \multicolumn{1}{c}{True rate} \\
\midrule
Dev   & 2,956  & 3,107  & 6,063  & 48.8\% \\
Test  & 3,073  & 2,872  & 5,945  & 51.7\% \\
Train & 24,391 & 23,298 & 47,689 & 51.1\% \\
\midrule
All   & 30,420 & 29,277 & 59,697 & 51.0\% \\
\bottomrule
\end{tabular}

\end{table}

\begin{table}[ht]
\caption{Label (accepted or not) distribution of claims by parties.\label{tab:stat_claims_by_parties}}
\centering
\small
\begin{tabular}{@{}lrrrrrrrr@{}}
\toprule
      & \multicolumn{4}{c}{Plaintiff}     & \multicolumn{4}{c}{Defendant}     \\
\cmidrule(rl){2-5}
\cmidrule(rl){6-9}
 &
  \multicolumn{1}{c}{True} &
  \multicolumn{1}{c}{False} &
  \multicolumn{1}{c}{All} &
  \multicolumn{1}{c}{True rate} &
  \multicolumn{1}{c}{True} &
  \multicolumn{1}{c}{False} &
  \multicolumn{1}{c}{All} &
  \multicolumn{1}{c}{True rate} \\
\midrule
Dev   & 1,494  & 1,530  & 3,024  & 49.4\% & 1,462  & 1,577  & 3,039  & 48.1\% \\
Test  & 1,765  & 1,342  & 3,107  & 56.8\% & 1,308  & 1,530  & 2,838  & 46.1\% \\
Train & 12,960 & 12,271 & 25,231 & 51.4\% & 11,431 & 11,027 & 22,458 & 50.9\% \\
\midrule
All   & 16,219 & 15,143 & 31,362 & 51.7\% & 14,201 & 14,134 & 28,335 & 50.1\% \\
\bottomrule
\end{tabular}
\end{table}

\begin{table}[ht]
\caption{Length statistics of claims and undisputed facts. Mean, standard deviation, and percentiles.\label{tab:stat_length}}
\centering
\small
\begin{tabular}{@{}lrrr@{}}
\toprule
      & \multicolumn{2}{c}{CL}                                        & \multicolumn{1}{c}{UF} \\
\cmidrule(rl){2-3}
      & \multicolumn{1}{c}{Plaintiff} & \multicolumn{1}{c}{Defendant} & \multicolumn{1}{c}{}   \\
\midrule
mean  & 120.8                         & 111.8                         & 129.5                  \\
std   & 219.6                         & 178.2                         & 171.7                  \\
min   & 2.0                           & 2.0                           & 2.0                    \\
25\%  & 63.0                          & 59.0                          & 59.0                   \\
50\%  & 94.0                          & 89.0                          & 94.0                   \\
75\%  & 140.0                         & 133.0                         & 151.0                  \\
99\%  & 501.2                         & 417.0                         & 707.0                  \\
max   & 19893.0                       & 16413.0                       & 7835.0\\
\bottomrule
\end{tabular}
\end{table}

Table~\ref{tab:dataset_overview} shows an overview of JTD. 
3,477 documents were annotated, which resulted in 7,978 instances. 
39.8\% of the instances are labelled as True for the $T$ (Table~\ref{tab:stat_tort}).
Table~\ref{tab:stat_claims} shows basic statistics for Claims. In total, 59,697 Claims are available for the rationale extraction task.
51.0\% of Claims are labelled as accepted (True), and others as rejected (False).
Table~\ref{tab:stat_claims_by_parties} gives numbers according to parties (plaintiff or defendant).
The rate of true-labelled Claims is consistent across parties, at 51.7\% for the plaintiff and 50.1\% for the defendant.
Out of 59,697 Claims, 52.5\% are from the plaintiff side, and others come from the defendant side.
Table~\ref{tab:stat_length} summarises statistics on the length of Claims and Undisputed Facts.
The maximum length of Claims from plaintiff and defendant can be outliers given that the 99th percentiles are much lower than them.
Table~\ref{tab:data_split} shows the split of our dataset.

\section{Models}
\begin{figure}[t]
\centering
\includegraphics[width=1.0\linewidth]{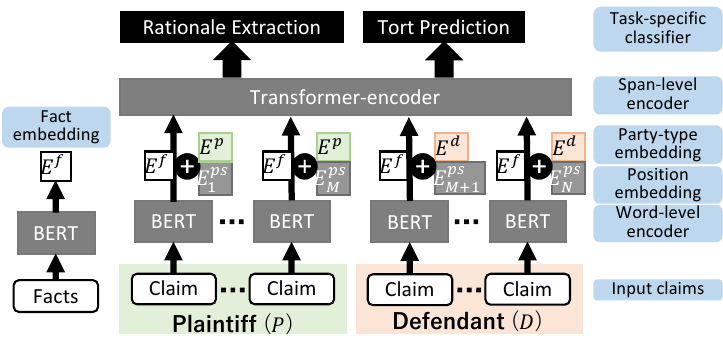}
\caption{Architecture of the IST models for legal judgment prediction with rationale extraction\label{fig:model_architecture}, where there are $N$ claims in total and $M$ out of $N$ claims are from Plaintiff. Fact embedding $E^{f}$ is shared across all claims, while party-type embeddings $E^{p}$ and $E^{d}$ are exclusive to the plaintiff's and defendant's claims, respectively. $E^{ps}_{n}$ is a position embedding for the $n$-th claim.}
\end{figure}
To establish an LJP baseline using JTD, we employ a hierarchical Transformer architecture, which achieves competitive performance in various legal NLP tasks~\cite{DBLP:journals/corr/abs-2110-00976}.
Our hierarchical Transformer-based models (Figure~\ref{fig:model_architecture}) are designed to capture both word-level context and span-level context by implementing a span-level Transformer encoder on top of word-level Transformer encoders. 
We call this model Inter-Span Transformer~(IST).
IST takes a sequence of spans as input, where each span corresponds to a claim from the plaintiff or defendant side. 
Its outputs are a single Boolean flag for tort judgment prediction and a sequence of Boolean flags for rationale extraction. 
These final outputs are obtained through a linear layer just after the output from the span-level encoder.
 IST accepts party-type embeddings to distinguish between the plaintiff's claims and the defendant's claims by assigning different embeddings according to who submitted the claim.
IST also considers positions of claims in inputs via position embeddings.
As an auxiliary input, IST takes undisputed facts.
To inject features from undisputed facts into every encoded claim, IST independently encodes all the concatenated undisputed facts into a single vector, namely fact embeddings.
An input representation to the span-level encoder, which corresponds to $n$-th claim by a plaintiff, is the sum of a claim vector from the word-level encoder, fact embeddings ($E^{f}$), party-type embeddings of the plaintiff ($E^{p}$), and position embeddings ($E^{ps}_n$).

As word-level encoders, we utilise two different types of pretrained BERT~\citep{devlin-etal-2019-bert} models, BERT-base-Japanese (BERTja)\footnote{\url{https://huggingface.co/cl-tohoku/bert-base-japanese-v2}} and Japanese-LegalBERT (JLBERT)~\citep{miyazaki-etal-2022-cross}.
Both BERTja and JLBERT use the same model architecture as the original BERT-base model with 12 layers, 768 dimensions of hidden states, and 12 attention heads.
We feed a vector corresponding to the [CLS] token from BERT output into a linear layer and use its output as a claim vector.
While BERTja is pretrained only with the Japanese Wikipedia corpus, JLBERT is adapted to the Japanese legal domain. JLBERT is firstly pretrained with the Japanese Wikipedia corpus and then further pretrained with Japanese judgment documents in Civil code cases.

Our JTD also uses Japanese judgment documents as the source of the dataset, and we manually checked the number of overlapping documents between the JLBERT pre-training data and the JTD test set.
We found 97.4 \% of documents from the JTD test in the pre-training data as well.
Nevertheless, we decided to employ JLBERT in our experiments because of the following three reasons: 1)~JLBERT is the only available pretrained language model in the Japanese legal domain, 2)~JLBERT was pretrained on the proprietary judgment document corpus so that we cannot pre-train our version of JLBERT model pretrained without documents in the JTD test set.
3)~As our dataset and tasks are produced with manually annotated judgment documents, masked language modelling cannot steal a look into our two LJP tasks.
Our JTD test set is still unseen data for BERTja-based models.
We can perform a sanity check by conducting the same experiments with BERTja. 
By comparing BERTja-based and JLBERT-based models, we avoid overestimating the performance of JLBERT-based models and provide an informative oversight of how the domain-adopted models behave in the Japanese legal judgment tasks.

\subsection{Rationale Extraction (RE)}
Rationale Extraction task identifies accepted arguments in the parties' arguments.
Inputs are undisputed facts ($U$) and arguments from both parties ($P$ from plaintiffs and $D$ from defendants).
Outputs are two sequences of Boolean values, $R^P$ for plaintiffs and $R^D$ for defendants, denoting accepted arguments as True.

The models used for RE are as follows.
\begin{itemize}[leftmargin=*]
    \setlength{\itemsep}{0cm} 
    \item \textbf{RE-random}: A random baseline that produces prediction based on the ratio of labels in the training set.
    \item RE-BERT: A BERT-based binary classifier. It classifies each input claim independently without regarding context across claims. There are two variants: \textbf{RE-BERTja} and \textbf{RE-JLBERT} depending on the BERT model used.
    \item RE-IST: An IST-based classifier. It outputs a sequence of Boolean values for RE, taking a sequence of claims from plaintiffs and defendants as inputs. There are \textbf{RE-IST-BERTja} and \textbf{RE-IST-JLBERT}.
\end{itemize}

\subsection{Tort Prediction (TP)}
Tort Prediction predicts whether a tort is affirmed ($T$, a Boolean value), given undisputed facts ($U$) and arguments from both parties ($P$ and $D$) as inputs.

The following are models used for TP.
\begin{itemize}[leftmargin=*]
    \setlength{\itemsep}{0cm} 
    \item \textbf{TP-random}: A random baseline, similar to RE-random.
    \item \textbf{TP-RF-meta}: A randomforest~\citep{DBLP:journals/ml/Breiman01} classifier. This model takes non-textual features as inputs: the year of a case, the court to which a case belongs and the number of claims from each party\footnote{We actually distinguish Factual Claims and Claims of Norms for the features.}. This baseline shows how well a model can predict court outcomes using meta-level features.
    \item \textbf{TP-RF-gold}: A similar model to TP-RF-meta, but this model uses the acceptance rates of each claim type instead of their number. The acceptance rates are calculated according to the gold labels of the rationale extraction task in JTD.
    Note that the target task here is TP, and this model is still blind against the gold labels of TP during validation and testing. This baseline provides a milestone for the TP task.
    \item TP-RF-cascaded: A classifier similar to TP-RF-gold. It uses the predicted rationales instead of the golds to calculate acceptance rates.
    \item TP-IST: An IST-based classifier. It outputs a Boolean value for TP, taking all claims from plaintiffs and defendants as inputs. This model also has two variants: \textbf{TP-IST-BERTja} and \textbf{TP-IST-JLBERT}.
    \item TP-IST-gold: An IST-based classifier whose input is only accepted gold claims. 
    This is an IST version of TP-RF-gold. There are \textbf{TP-IST-gold-BERTja} and \textbf{TP-IST-gold-JLBERT}.
    \item TP-IST-cascaded: Its architecture is identical to TP-IST-gold. This model takes only predicted rationales by RE as inputs.
\end{itemize}

\subsection{Multi-task Approach}
We also take a multi-task approach with the IST architecture, in which the model learns its parameters jointly for both RE and TP tasks using the combined loss function~(\ref{eq:loss_multi}).

\begin{equation}
\label{eq:loss_multi}
\mathrm{Loss} = {\alpha}\mathrm{Loss}_\mathrm{TP} + {(1 - \alpha)}\mathrm{Loss}_\mathrm{RE}
\end{equation}
The multi-task IST takes a sequence of claims and outputs a sequence of Boolean predictions for RE and a Boolean value for TP.
There are two variants, \textbf{Multi-IST-BERTja} and \textbf{Multi-IST-JLBERT}.

\section{Experiments}
\subsection{Experimental Settings}
In the training phase of all neural network-based models, we used the AdamW~\citep{DBLP:conf/iclr/LoshchilovH19} optimiser with a linear scheduler whose warmup step was 10\% of total training steps. Epochs were fixed at 30. 
The maximum length of the word-level encoders is 512. IST models accept up to 64 claims.
We chose the best-performing checkpoint according to accuracy scores in the development set.
The parameters of word-level encoders (BERTja and JLBERT) are not frozen and fine-tuned.

We employed an optimisation framework Optuna~\citep{optuna_2019} to search optimal hyperparameters for each neural network-based model.
All hyperparameters were tuned using 3,000 instances of the training data.
For the RE-BERT models, we performed a grid search to find the optimal learning rate where the total number of trials was four.
For the IST models, we conducted more extensive searches. 
We utilised the Tree-structured Parzen Estimator algorithm~\citep{DBLP:conf/nips/BergstraBBK11} for IST as their search spaces were much larger than RE-BERT's. The total number of trials was 112.
Table~\ref{tab:hyperparameter} shows their search spaces.
The ``TRenc'' hyperparameters are for the Transformer module implemented as the span-level encoder in the IST.
``Use UF'' is a flag whether a model utilises fact embeddings or not.
$\alpha$ is a weight of the loss function (Eq~(\ref{eq:loss_multi})) for the multi-task models.

\begin{table}[t]
\caption{Hyperparameters search space\label{tab:hyperparameter}}
\centering
\small
\begin{tabular}{lr}
\toprule
Parameters                              & \multicolumn{1}{c}{Choices}\\
\midrule
Learning rate                 & 2e-6, 4e-6, 6e-6, 8e-6 \\
TRenc heads                 & 2, 4, 6, 8 \\
TRenc FF dim                 & 64, 128, 256, 512 \\
TRenc layers & 1, 2, 3, 4 \\
Use UF & True, False \\
$\alpha$ (if applicable) & 0, 0.05, ..., 0.95\\
\bottomrule
\end{tabular}
\end{table}

We adopt accuracy for the evaluation metrics for both RE and TP.
We trained and tested each model five times with different random seeds and averaged the scores.
We also performed the permutation test to assess the statistical significance between models (significance level is at $p<0.05$, two-tailed test).
The target metric of the permutation test is the accuracy score.

\subsubsection{Cascaded Model Settings}
In the cascaded models, an RE model first predicts claims to be accepted, which are then fed to a TP model.
We employ the outputs from the best-performing RE model.
To be concrete, we chose Multi-IST-JLBERT and Multi-IST-BERTja for the RE task in the cascaded model.
We notate the cascaded models using these RE models as follows: \textbf{TP-RF-cascaded-BERTja}, \textbf{TP-RF-cascaded-JLBERT}, \textbf{TP-IST-cascaded-BERTja}, and \textbf{TP-IST-cascaded-JLBERT}.

\subsection{Results}
\begin{table}[t]
\caption{Experimental results of RE (accuracy with standard deviation)\label{tab:rationale_extraction_results}.}
\centering
\small
\begin{tabular}{@{}llrrr@{}}
\toprule
               & \multicolumn{3}{c}{\textbf{Claim-level}}       & \multicolumn{1}{c}{\textbf{Doc-level}} \\ \cmidrule(lr){2-4}
\textbf{Models}  & \multicolumn{1}{c}{\textbf{All}}              & \multicolumn{1}{c}{\textbf{Plaintiff}} & \multicolumn{1}{c}{\textbf{Defendant}} & \multicolumn{1}{l}{} \\ \midrule
RE-random      & 0.498 (.005) & 0.496 (.005) & 0.501 (.007) & 0.502 (.007) \\
RE-BERT-BERTja & 0.598 (.005) & 0.597 (.007) & 0.598 (.008) & 0.622 (.003) \\
RE-BERT-JLBERT & 0.634 (.005) & 0.635 (.009) & 0.631 (.006) & 0.653 (.006) \\
RE-IST-BERTja  & 0.637 (.012) & 0.652 (.010) & 0.620 (.022)& 0.658 (.014) \\
RE-IST-JLBERT  & 0.663 (.008) & \textbf{0.677} (.013) & 0.648 (.008) & 0.681 (.007) \\
Multi-IST-BERTja & 0.666 (.008) & 0.671 (.009) & 0.661 (.011) & 0.690 (.013) \\
Multi-IST-JLBERT & \textbf{0.674} (.009) & 0.675 (.007) & \textbf{0.673} (.014) & \textbf{0.691} (.005) \\ \bottomrule
\end{tabular}
\end{table}

\begin{table}[t]
\caption{Experimental results of TP (accuracy with standard deviation)\label{tab:tort_results}.}
\centering
\small
\begin{tabular}{@{}lr@{}}
\toprule
\textbf{Models} & \textbf{Macro avg.} ($\sigma$) \\ \midrule
TP-RF-gold                      &  0.880 (.001) \\ 
TP-IST-gold-BERTja                       &    0.883 (.005) \\ 
TP-IST-gold-JLBERT                   &    0.883 (.009) \\ \midrule
TP-random                                &   0.503 (.014)\\ 
TP-RF-meta                     & 0.574 (.006)\\ 
TP-IST-BERTja                             &   0.649 (.023)\\ 
TP-IST-JLBERT                         &    0.674 (.024)\\ 
Multi-IST-BERTja                          &   0.680 (.007)\\ 
Multi-IST-JLBERT                      &    \textbf{0.683} (.020)\\ \midrule
TP-RF-cascaded-BERTja                  &  0.660 (.030)\\ 
TP-RF-cascaded-JLBERT              &  0.639 (.012)\\ 
TP-IST-cascaded-BERTja         &    0.673 (.022)\\ 
TP-IST-cascaded-JLBERT &    0.666 (.013)\\ \bottomrule
\end{tabular}
\end{table}

\subsubsection{Rationale Extraction}
Table~\ref {tab:rationale_extraction_results} shows experimental results for the RE task. The ``All'' column shows accuracy scores calculated on all claims. We use ``All'' scores as the target metric for the permutation test.
The ``Plaintiff'' and ``Defendant'' columns show accuracy scores of claims from plaintiff and defendant, respectively.
The ``Doc-level'' column shows accuracy scores at the document level, where the scores are first calculated per document and then averaged over documents.

According to ``All'' scores, IST showed significant improvement from RE-BERT; capturing the context between input claims helped the task.
Comparing BERTja and JLBERT, we find the model using JLBERT always performs numerically better than that with BERTja, confirming the statistically significant difference except for pairs of multi-task approaches.
The multi-task models (Multi-IST-$*$) always showed better accuracy than their corresponding single-task models (RE-IST-$*$).
The improvement by the multi-task model is more significant when using BERTja than JLBERT.
The improvement from RE-IST-BERTja to Multi-IST-BERTja is statistically significant.

When we see scores according to parties, the overall trend is the same as ``All'' scores. An exception is ``Plaintiff'' where RE-IST-JLBERT achieved the best score among all the models.

The ``Doc-level'' scores share the same trend with ``All''.

\subsubsection{Tort Prediction}
Table~\ref {tab:tort_results} shows experimental results for the TP task.
All TP-IST and Multi-IST models were significantly better than TP-RF-meta in accuracy.
Gold models ($*$-gold-$*$) showed approximately 0.88 in accuracy, suggesting an upper bound of the TP models given the perfect RE output. 

The multi-task models showed the best performance in TP as well as RE.
The multi-task models are always numerically better than their corresponding single-task models in accuracy, and we observed statistical significance between Multi-IST-BERTja and TP-IST-BERTja.

The cascaded models showed improvement from their corresponding single-task models; however, even the best model (TP-IST-cascaded-BERTja) did not come close to the Multi-IST models. 

\subsection{Discussion}

In both tasks, the multi-task models show the best performance, suggesting they could leverage interaction between the tort judgment and its supporting arguments.
This result is aligned with legal experts' behaviour in interpreting legal cases. 
They do not simply conclude in a bottom-up manner.
Rather, they make inferences moving between consideration of subordinate arguments and the conclusion. 
We assume the multi-task models that jointly learned both TP and RE could model the legal prediction better.

The best accuracy we achieved in the TP task was 0.683 with JLBERT as a word-level encoder and 0.680 with BERTja.
Those scores are far from perfect and still much lower than 0.880, which is the milestone achieved by the gold models.
Our scores are comparable with an accuracy score of 0.668 reported in the class action LJP task from the United States~\citep{semo-etal-2022-classactionprediction}\footnote{\citet{semo-etal-2022-classactionprediction} works on similar tasks to our TP task. However, many differences still exist, such as languages, jurisdictions, target topics, and legal proceedings. They may affect the outcome of experiments. Thus, this comparison is only for advisory.}.

The high accuracy of the gold TP models indicates the importance of RE for TP. 
The best RE accuracy was 0.674 with JLBERT and 0.666 with BERTja. There is a big room to be improved from those baseline models yet.
JTD implements claim-level rationale extraction for LJP while previous work employed paragraph-level~\citep{Chalkidis2021-bh}.
Challenges of JTD stem from finer-grained targets to be classified where more precise inference is required.

In the RE task, we observed that IST models ($*$-IST-$*$) showed higher accuracy for the plaintiff's claims than the defendant's claim.
Differences in the number of instances between the plaintiff and the defendant could have caused this result. JTD has more claims from the plaintiff (52.5\%) than the defendant (47.5\%) (Table~\ref{tab:stat_claims_by_parties}). 

JLBERT is generally superior to BERTja for the same architecture for both tasks except for the cascaded models.
However, the difference is small for the best models (Multi-IST-$*$).
As our tasks are specific to the legal domain and JLBERT was trained by the judgment documents that are the source of JTD, we expected better performance from the JLBERT-based models. In reality, however, the effectiveness of JLBERT was limited.
This indicates that the span-level encoder and its following layers play more important roles than the word-level encoders in our tasks.

\section{Error Analysis}
We conducted a detailed error analysis employing human legal experts on the tort prediction outputs to identify the source of prediction errors.

\subsection{Analysis Setup}
Four experts (the authors of this paper) participated in the analysis.
They are all legal professionals, including three professors of Law at Japanese universities and a Japanese lawyer who has eight years of experience.

We consider the outputs from the well-performed four models: TP-IST-BERTja, TP-IST-JLBERT, Multi-IST-BERTja and Multi-IST-JLBERT.
We merged the TP outputs by majority voting from five runs of each model.
We analysed instances where all four models failed in prediction.
We obtained 139 out of 811 instances from our test set.
Due to limited human resources, we randomly selected 60 from 139 instances.
Two of the experts got 20 instances each, and the other two got 10 each.
The experts were asked to fill out a form given the same inputs as the models.
They were allowed to see the original judgment documents, if necessary.
There are two major questions in the form.

\subsubsection{Human Confidence Score} The experts were asked to rate how confidently they could make tort predictions with the given input.
The score scale is 0, 1, 2, 3, where 3 means a human legal expert can confidently judge whether an instance is a tort or not, 2 means a human legal expert can predict with uncertainty, 1 means a human legal expert can only predict a tendency of its outcome and 0 means even a human legal expert cannot predict its outcome at all.

\subsubsection{External Knowledge} We hypothesised that a certain type of error occurred because of missing necessary information in the input texts,
i.e., legal domain-specific knowledge and facts described in an external document other than a judgment document.
We asked the experts to identify what kind of external knowledge was necessary to make a correct prediction.
There are four options, ``General knowledge'', ``Legal knowledge'', ``Insufficient input'', and ``Other''.
``Insufficient input'' is chosen when an expert believes essential information specific to the instance is missing from input claims or undisputed facts, while ``General knowledge'' and ''Legal knowledge'' would be chosen when missing information is independent of the instance.
``Other'' is chosen when the above three do not fit.
The necessary information is detailed for the ``Other'' option in the free text.
The experts could choose multiple options from the four.

In addition to the two questions above, the experts could submit their feedback and comments.

\subsection{Result\label{sec:humananalysis_result}}
\begin{figure}[t]
\centering
\includegraphics[width=0.8\linewidth]{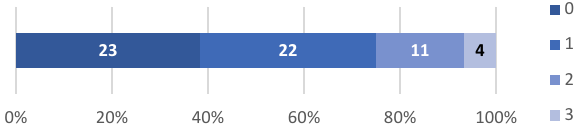}
\caption{Confidence scores by human expert}
\label{fig:confidence_score}
\end{figure}

Figure~\ref{fig:confidence_score} shows human confidence scores on 60 instances.
There are 23 instances in which even human experts cannot predict their outcomes at all, while the experts make only four confident predictions.
Scores 0 and 1 make more than 70\% of the analysed instances, which is considered unconfident.
The results suggest those wrongly predicted instances by models are also difficult for a human legal expert.

\begin{table}[hptb]
    \caption{Missing information for the TP task}
    \centering\small
    \begin{tabular}{lcr}
         \toprule
         Missing information& \#instances& [\%]\\
         \midrule
         General knowledge& \phantom{0}2& 3.3\\
         Legal knowledge& 15& 25.0\\
         Insufficient inputs& 35& 58.3\\
         Other& 17& 28.3\\
         \bottomrule
    \end{tabular}
    \label{tab:missing_knowledge}
\end{table}

52 out of 60 instances are flagged with one or more types of external knowledge.
Table~\ref{tab:missing_knowledge} shows the distribution of instances across the missing information types.
Only two instances require ``General knowledge''.
One of them is about a sender's information disclosure, which requires missing information ``the same IP address does not necessarily mean that the same person posted the messages''.

The experts found 15 instances require ``Legal knowledge'' to predict their outcome correctly. The legal knowledge includes specific requirements of certain laws, civil code procedures, and heuristics knowledge, which a legal expert can obtain through their experience.

The dominant category was ``Insufficient inputs'' (35 instances), which
stem from the nature of the judgment document format.
Judgment documents often refer to external documents for detailed evidence.  In such cases, descriptions and claims about the evidence in the judgment document itself may not be specific. Often, those external documents are not publicly available.
Thus, the annotated claims and undisputed facts from the judgment document might be insufficient for confident prediction.
Also, in some judgments, concrete descriptions of evidence or even factual claims from the parties can be pseudonymised or censored to protect private and confidential information.

Another cause of insufficient inputs comes from the annotation process. For example, an annotator failed to extract necessary claims and facts during the annotation process.
Our annotation guideline prohibits, in principle, the extraction of claims and facts from the ``court's decision'' section, which contains information corresponding to the answer in the TP task~(\ref{sec:dataset_source}).
Therefore, the necessary facts were not fully annotated when they were described only in the ``court's decision'' section.

Eight out of 60 instances are not flagged with any external knowledge. However, five are rated with a confidence score of 1, suggesting that they are not necessarily easy to predict.
The experts reported that those instances were from controversial cases, including a case whose decision was flipped in a higher court. Such a case is considered challenging for a machine.

Many comments and feedback from the experts are on instances extracted from complicated cases with multiple tortious actions to be judged.
Our annotation guideline instructs annotators to distinguish one tortious action from others.
When there are many tortious actions in the same documents, there are many related claims and facts, so argumentative relations between them are complicated. 
For example, a single claim can be related to multiple tortious actions.
Thus, an annotator can miss necessary facts or fail to exclude unnecessary claims in the instance, resulting in instances with insufficient inputs. Moreover, even if instances in such cases are annotated perfectly,  they are still challenging to predict their outcomes because of the tangled argumentative relations between claims.

Counterclaims make instances more complicated, which allows defendants to assert a new claim against plaintiffs during the same lawsuit that the plaintiffs initially filed. The experts identified three out of 60 instances where counter-claims made the instances difficult.
In annotating counterclaim instances, our annotation guideline explicitly instructed annotators to extract the original plaintiffs' claims as defendant claims and the original defendants' claims as plaintiff claims to make argumentative relations between claims consistent with non-counterclaim instances.
Nevertheless, we found the counterclaim instances still confusing to a machine.
For example, the counterclaim's plaintiff was expressed with ``defendant'' in the input and vice versa.
Although these notations were correct for a counterclaim instance, they can confuse a machine since most instances are non-counterclaim.

Another remarkable feedback is about instances from medical cases.
In those cases, detailed medical procedures and expert opinions, which are crucial to making a legal decision, may be prepared separately from the main judgment documents.
As a result, our annotators could not extract the necessary information for such instances. Therefore, these medical cases can be difficult to predict.

\section{Limitations and Ethical Considerations}
We clarify limitations and ethical considerations here to prevent misuse of our dataset and misunderstanding of our findings.

The task design for LJP reflects essential elements of the Japanese jurisdiction. However, there are differences from real-world conditions. The inputs in our tasks are only Undisputed Facts, Plaintiff's Claims, and Defendant's Claims, which are obtained from judgment documents. In real court cases, there are other documents, including third-party expert opinion letters, detailed evidence documents, and other undisclosed documents (e.g., private information and confidential patent information). They are intermediate outputs or auxiliary inputs in court cases but are still important. This difference limits the model's capability, as shown by manual error analysis results (\ref{sec:humananalysis_result}). The difference often stems from the limited scope of publicly available information. We suspect such limitation also applies to other jurisdictions that do not disclose the whole court document sets. As many LJP datasets are constructed from only judgment documents, we must recognise that the current LJP task design can be limited and reflect only a part of legal decision-making.

The annotation scheme for our dataset was validated through the preliminary experiment with 25 documents annotated by five legal experts. We note that the number of documents used in the preliminary experiment is lower than that of our final dataset, i.e. 3,477 documents, and we did not perform dual-coder annotation for the final dataset. Although we took alternative measures (extended tutorials, and annotator selection via dry-run) to ensure the quality of annotation, we acknowledge that the final dataset may contain inevitable human errors.

We also acknowledge that our dataset is limited in its scope. We must point out that the dataset contains tort-related cases, which is a part of the whole legal judgement in the Japanese Civil jurisdiction. Thus, our findings from the experimental results cannot directly apply to the general Japanese LJP tasks. However, we believe our baseline results provide informative clues to other civil legal judgement tasks. Another limitation is the quality of the annotations. While we have made an effort to keep the annotation quality as high as possible, there is still a chance of errors and misunderstandings by the annotators. Although the agreement study shows reasonable performance, we plan to update and expand our JTD on a long-term schedule continuously.

Given the limitations, we emphasise that one should not rely solely on a model trained on JTD in one's legal decision-making.
Our recommended use case of the JTD-trained models is legal assistance services, but they should not be fully automated. We recommend using the models with legal professionals such as lawyers\footnote{As of the paper submission, non-lawyers are prohibited from engaging in any legal services for the purpose of earning compensation, which are exclusively licensed lawyers in Japan. (\href{https://www.japaneselawtranslation.go.jp/en/laws/view/3636\#je_ch10at1}{Attorneys Act, Art.~72}) }. This human-machine hybrid approach will improve the lead time of case handling and correct potential errors from models' false predictions.

Moreover, we emphasise that JTD is not intended to develop a judgment system. In other words, we did not intend to replace judges or courts with JTD-trained models.
Important intellectual activities of judges include the conceptualisation of new rights and the update of law interpretations in response to age. However, the current LJP tasks in JTD do not cover such aspects. Therefore, we argue only human judges should take responsibility for such authority.

\section{Conclusion and Future Work}
We proposed a novel dataset for the Japanese legal judgment prediction featuring tort cases under the Japanese Civil Code.
We specifically targeted two tasks: tort prediction and rational extraction.
This is the first dataset consisting of real Japanese court cases with human expert annotation.
Although we still have to continue to improve the annotation quality, our annotation procedure will be a good starting point to produce reliable datasets in the Japanese legal domain.

We conducted a feasibility study of the two tasks with the baseline models. We also compared the performance between single-task and multi-task approaches.
The baseline experiments confirmed the feasibility of our tasks, and we found that the multi-task approach performed better.
Moreover, we manually analysed the outputs from the IST models with experienced legal specialists to diagnose the errors. 
The results suggest that there are difficult cases even for human experts.

Our dataset and extended expert analysis only made these novel findings possible.
We believe our dataset provides a useful resource in legal NLP for not just Japanese LJP research but also comparative experiments and analysis across different languages and jurisdictions.

In future work, we plan to extend the size of our dataset and increase the types of cases in addition to tort cases. Additionally, adapting our annotation scheme and tasks to other jurisdictions is an interesting direction for further research.

\backmatter
\bmhead{Acknowledgments}
We appreciate Prof. Souichirou Kozuka at Gakushuin University and Prof. Kazuhiko Yamamoto at Hitotsubashi University for their helpful comments.

\bmhead{Declarations}
The judgment documents data for this study was provided at no cost by LIC Co., Ltd. solely for academic research purposes.
This work was supported by JST RISTEX Grant Number JPMJRX19H3, JST ACT-X Grant Number JPMJAX20AM and Support Centre for Advanced Telecommunication Technology Research.

\bmhead{Supplementary information}
Our JTD dataset is available for non-commercial academic research purposes.
We also share the original document set used in JTD construction.
All the files are available as ``Japanese Tort-case Dataset (日本語不法行為事件データセット)'' at \url{https://www.gsk.or.jp/catalog/}.

\clearpage
\begin{appendices}




\section{Prediction Examples}
Table~\ref{tab:hardcasesample} shows a prediction example.
Translations are ours. A sentence (claim) in Japanese might be separated into several sentences in English as a result of translation. The sample was chosen from instances in which a human expert commented that a judicial decision prediction would be difficult.

\begin{table}[h]
\caption{Prediction sample (ID: L07530809-6654).}
\centering
\small
\begin{tabular}{@{}lp{0.64\linewidth}ccl@{}}
\toprule
 &
   &
  \multicolumn{2}{c}{\textbf{RE labels}} &
   \\
\textbf{Party} &
  \textbf{Claim} &
  \multicolumn{1}{l}{\textbf{Gold}} &
  \multicolumn{1}{l}{\textbf{Pred.}} &
   \\ \midrule
Plaintiff &
  The defendants requested a loan from the plaintiff around July 2018, stating that ``the development of an accounting system using AI is close to being completed, but we do not have enough funds''.
  In response to this request, the plaintiff concluded the agreement on 30th July 2018 and granted a total of JPY 10 million on three occasions between 30th July and 25th September 2018. However, the defendants had not developed the AI accounting software at all. &
  True &
  False &
   \\ &&&\\
Plaintiff &
  The defendants did not develop any AI accounting software. Despite having no intention of using the funds borrowed from the plaintiff for the defendant company's business or repaying them, they pretended as if they had such intentions and fraudulently obtained money from the plaintiff. Such fraudulent acts by the defendants constitute tortious conduct against the plaintiff.&
  True &
  False &
   \\ &&&\\
Defendant &
  There is no fact that the defendants asked the plaintiff for the loan, and the agreement was initially triggered by the plaintiff's offer to purchase shares of the defendant company for 20 million yen as the plaintiff had made several hundred million yen in profits from the plaintiff's business. Subsequently, due to the plaintiff's circumstances, the plaintiff agreed to lend the defendant 10 million yen. &
  False &
  True &
   \\ &&&\\
Defendant &
  The defendants are developing AI accounting software, and moreover, under the agreement, the use of the loans was only specified as ``operating funds'' and not limited to the cost of developing AI accounting software. Thus, there was neither a breach of the agreement by the defendants nor tortious conduct. &
  False &
  True &
   \\ \midrule
   \multicolumn{4}{c}{\textbf{TP Gold}: True, \textbf{TP Prediction}: False}
   \\ \bottomrule
\end{tabular}
\label{tab:hardcasesample}
\end{table}

\end{appendices}
\clearpage


\bibliography{sn-bibliography}

\end{document}